# Fast tree skeleton extraction using voxel thinning based on tree point cloud

**Author:** Jingqian Sun[1], Pei Wang[1*], Ronghao Li[1], Mei Zhou[2]

**Address:** [1]School of Science, Beijing Forestry University, No.35 Qinghua East Road, Haidian District, Beijing 100083, China

[2]Key Laboratory of Quantitative Remote Sensing Information Technology, Aerospace Information Research Institute, Chinese Academy of Sciences, Beijing 100094, China

**Corresponding author:** Pei Wang[1*]

**E-mail:** Jingqian Sun: sunjq_2019@bjfu.edu.cn

Pei Wang: wangpei@bjfu.edu.cn

lironghao@163.com

zhoumei@aoe.ac.cn

[*]**For corresponding E-mail:**

wangpei@bjfu.edu.cn

# Abstract

Tree skeleton plays an important role in tree structure analysis, forest inventory and ecosystem monitoring. However, it is a challenge to extract a skeleton from a tree point cloud with complex branches. In this paper, an automatic and fast tree skeleton extraction method (FTSEM) based on voxel thinning is proposed. In this method, a wood-leaf classification algorithm was introduced to filter leaf points for the reduction of the leaf interference on tree skeleton generation, tree voxel thinning was adopted to extract raw tree skeleton quickly, and a breakpoint connection algorithm was used to improve the skeleton connectivity and completeness. Experiments were carried out in Haidian Park, Beijing, in which 24 trees were scanned and processed to obtain tree skeletons. The graph search algorithm (GSA) is used to extract tree skeletons based on the same datasets. Compared with GSA method, the FTSEM method obtained more complete tree skeletons. And the time cost of the FTSEM method is evaluated using the runtime and time per million points (TPMP). The runtime of FTSEM is from 1.0 s to 13.0 s, and the runtime of GSA is from 6.4 s to 309.3 s. The average value of TPMP is 1.8 s for FTSEM, and 22.3 s for GSA respectively. The experimental results demonstrate that the proposed method is feasible, robust, and fast with a good potential on tree skeleton extraction.

# 1. introduction

The past decade of the terrestrial laser scanning (TLS) development has made the point cloud data an increasingly practical option for urban modeling (Zhang et al., 2018), object surface reconstruction (Berger et al., 2017), environmentology and ecosystems (Lamb et al., 2018), forestry (Kandare et al., 2017), forest filed inventories (Palace et al., 2016) and tree parameters estimation (Zheng et al., 2016).

Tree point cloud is a promising data to improve the tree parameters estimation, such as leaf area index (LAI) (Olsoy et al., 2016), tree crown volume (Kong et al., 2016), diameter at breast height (DBH) (Oveland et al., 2017), tree branch and stem biomass (Hauglin et al., 2013). Compared with the conventional forest inventory methods, TLS can fast provide accurate and reliable tree point clouds that offer high-precision and high-density without destroying trees.

Tree skeleton reflects not only the topological structure, but also the morphological structure which can be used to analyze the impact of environmental factors (Aiteanu and Klein, 2014a) and to reconstruct the 3D tree models (Bournez et al., 2017). The detailed and accurate information of the tree trunk and branches is reconstructed using the topology and geometric structure information of the tree skeleton, which can be used to estimate the order, relationship, length, angle, and volume information of the branches (Méndez et al., 2016; Zhang et al., 2020). Furthermore, the obtained information can also be used to study the biomass (Calders et al., 2015), precise tree pruning, tree growth assessment, and tree management

(Zhang et al., 2020). Tree point cloud has the advantage to describe the tree skeleton in detail due to the accurate and enormous data.

To extract tree skeletons from point clouds, some researchers have tested many novel methods with field experiments. Depending on the different processing purpose, these tree skeleton extraction methods can be grouped into two categories. In the first category, the tree skeleton extraction methods were used to reconstruct 3D trees based on tree point cloud. Gorte and Pfeifer (2004) used a 3D morphology method to segment and extract the skeleton from tree point cloud data. Xu et al. (2007) extracted tree skeletons using the shortest path algorithm and clustering algorithm. Yan et al. (2009) fulfilled tree skeletons using the combination of the k-means clustering method, cylinder detection method, and simple heuristics method. Based on the principle of graph structure, Livny et al. (2010) constructed a Branch-Structure Graph (BSG) using the Dijkstra's algorithm, and obtained a skeleton structure using the minimizing error function to refine the BSG. Aiteanu and Klein (2014a) used a hybrid method to extract tree skeletons based on the point density. For dense point cloud areas, the principal curvatures were used as an indicators for branches, detected ellipses in branch cross-sections and created branch skeletons; for sparse regions, spanning tree was used to approximate branch skeletons. Delagrange et al. (2014) developed an efficient method based on extensive modifications to the skeleton extraction method (Verroust and Lazarus 2000). Wang et al. (2014) generated tree skeletons using the distance minimum spanning tree (DMST) method. Hackenberg et al. (2015) obtained cylinder skeleton from tree point cloud using clustering method and cylinder fitting method. Wang et al. (2016) utilized the Laplacian contraction method to extract tree skeletons from the raw point cloud. Mei et al. (2017) proposed a L1-minimum spanning tree (L1-MST) algorithm to refine tree skeleton extraction, which integrates the advantages of both L1-median skeleton and MST algorithm. The methods in the first category are mainly aimed at 3D tree modeling and reconstruction, few trees were tested using each method.

Furthermore, in the second category, some works focused on extracting tree skeletons from tree point clouds directly. Bucksch and Lindenbergh (2008) proposed a Collapsing And Merging Procedures IN Octree-graphs (CAMPINO) method to extract the skeletons of two trees, which were an apple tree and a cherry tree. Whereas, the CAMPINO method is limited by the resolution of octree, and the topological correctness was not fully solved. Bucksch et al. (2010) also proposed a skeletonization method based on octree and neighborhood connectivity principle, which required a manual input parameter leading to a low level of automation. Su et al. (2011) extracted the curve skeletons of various tree models based on point cloud contraction using constrained Laplacian smoothing. There are the plum tree, apple tree and cherry tree tested using the method without time cost values. Bremer et al. (2013) generated tree skeleton using continuous iteration. Thirty-six simulated sample tree point clouds were analyzed to search for the best parameter setting. Li et al. (2017) extracted two tree skeletons of a toona tree and a peach tree based on k-means clustering method and breadth first search (BFS) method with more than 300 seconds. He et al. (2018) utilized the branch geometric features and local properties of point clouds to optimize tree skeleton extraction. Gao et al. (2019) reconstructed tree skeletons based on the Gauss clustering method and force field model, which firstly separates a tree into branches and leaves using the Gaussian mixture method and constructs tree skeleton only using the branches. Fu et al. (2020) extracted the initial skeleton from the tree point cloud using octree and level set methods, and optimized the skeleton using the cylindrical prior constraint (CPC) algorithm. Jiang et al. (2020) developed an iterative contraction method based on geodesic neighbors termed to extract tree skeletons. They dealt with a total of 10 artificially generated trees and 9 reconstructed trees, and the time cost of these trees ranged from 15.15 seconds to 82.47 seconds. However, the number of points in each tree point cloud was not stated. Some errors occur with a drastic branch angle change. Ai et al. (2020) presented an automatic tree skeleton extraction approach based on multi-view slicing, which borrowed the idea from the medical imaging technology of X-ray computed tomography. A leaf-on tree and a leaf-off tree were tested, and the time cost was 33 minutes, 18 minutes respectively.

In addition, some articles mainly realized the skeleton extraction of some general objects including few tree point clouds (Huang et al., 2013; Zhou et al., 2020). Some other articles proposed skeleton extraction methods to obtain skeletons from point cloud data of maize and sorghum (Wu et al., 2019; Xiang et al., 2019).

Although the above-mentioned methods have achieved satisfactory performance, they are limited on automation, accuracy, testing tree point cloud data and efficiency. First, some methods need manual intervention which reduce the performance of automation (Bucksch and Lindenbergh, 2008; Bucksch et al., 2010; Qin et al., 2020; Zhou et al., 2020). Second, some methods take measures to decrease the impact of noise, outliers and occlusions (Xu et al., 2007; Zhen Wang et al., 2014; Wang et al., 2016; Gao et al., 2019). Meanwhile, other methods have no strategy to improve the completeness of tree skeletons. Third, most methods were tested on few tree point clouds which should be verified using more data to confirm the robustness. Fourth, the time consumption of the above-mentioned methods needs to be evaluated, the reported time cost of the methods is tens of seconds, minutes or even hours even without mentioning the amount of point in the tree point clouds.

To improve tree skeleton extraction, this paper proposes a fast and automated approach which reduces the impact of leaf points using a wood-leaf classification, accelerates the process with a voxel thinning algorithm, and connects the broken branches using the branch positions and relationships.

This paper is organized as follows: Section 2 introduces the experimental data used in the paper, describes the method and explains the method in detail. Section 3 shows the skeleton extraction results of tree point clouds. Section 4 analyzes the skeleton results and discusses the advantages and limitations of the algorithm. Section 5 summarizes the characteristics of the proposed method and provides an outlook for the future work.

## 2. Materials and Methods

**2.1 Experimental data**

In this study, twenty-four willow trees (*Salix babylonica Linn and Salix matsudana Koidz*) from the Haidian Park, Haidian District, Beijing, China, were scanned using the RIEGL VZ-400 TLS scanner (RIEGL Laser Measurement Systems GmbH, 3580 Horn, Austria) which characteristics are shown in Table 1. In the scanning, the three dimensional information and intensity information are collected and recorded at the same time. The tree point clouds were manually extracted from the single-scan scene point clouds and the tree dataset was previously also used for the analysis of wood-leaf classification (Sun et al., 2021).

Table 1. The characteristics of RIEGL VZ-400 scanner.

| Technical parameters | |
| --- | --- |
| The farthest distance measurement | 600 m (natural object reflectivity ≥ 90%) |
| The scanning rate (points / second) | 300000 (emission), 125000 (reception) |
| The vertical scanning range | -40° ~ 60° |
| the horizontal scanning range | 0° ~ 360° |
| Laser divergence | 0.3 mrad |
| The scanning accuracy | 3 mm (single measurement), 2 mm (multiple measurements) |
| The angular resolution | better than 0.0005° (in both vertical and horizontal directions) |

## 2.2 Method

In this section, the proposed method will be systematically introduced, whose flowchart is shown in Figure 1. The proposed method mainly includes three parts, which are leaf points filtering, tree voxel thinning and tree skeleton building.

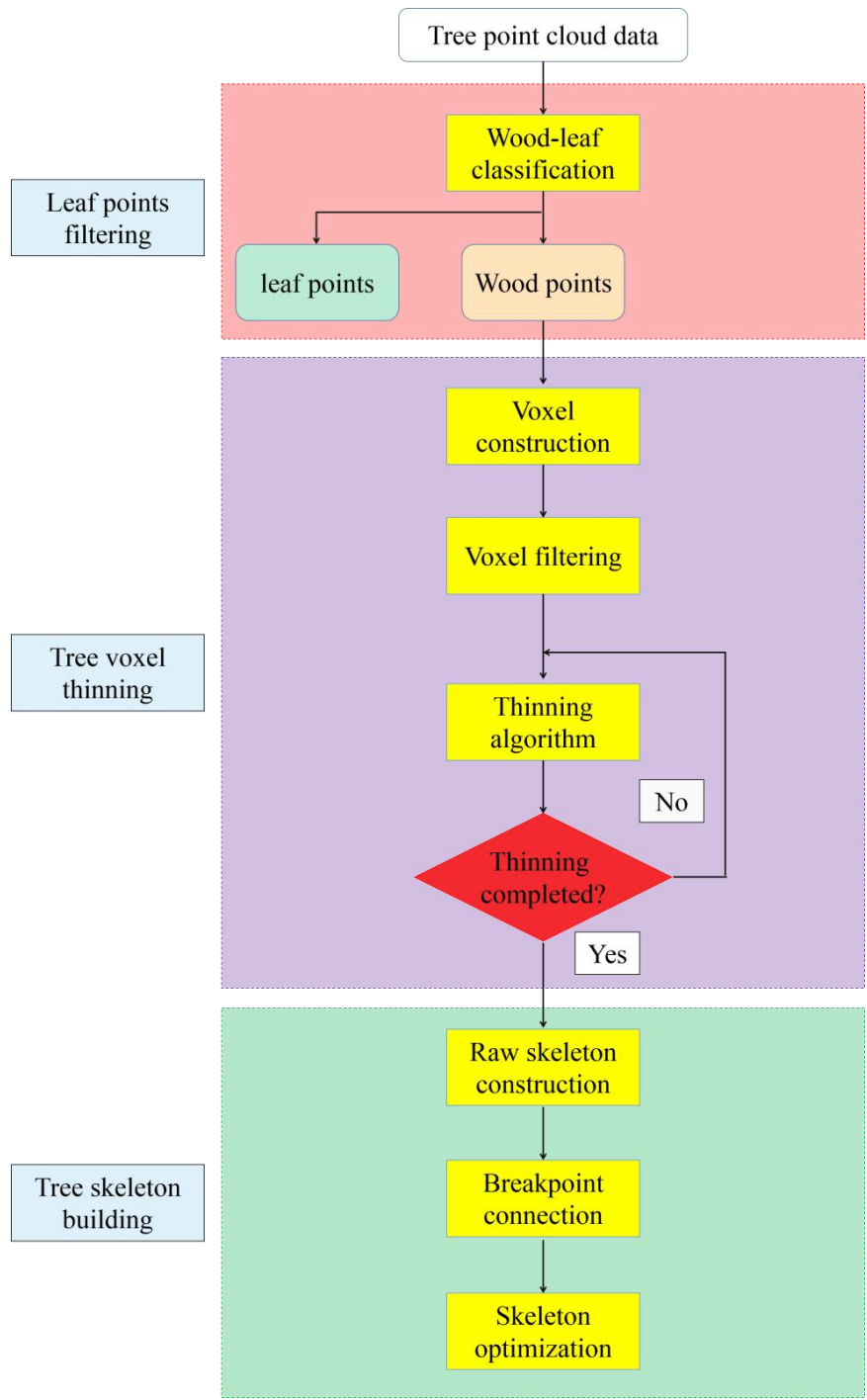

Fig. 1. Flow chart of FTSEM algorithm.

**2.2.1 Leaf points filtering**

It is well known that TLS instrument acquires tree point clouds using laser beams which cannot distinguish leaves and woody parts. The woody parts include trunk and branches which demonstrate the tree skeleton. However, leaf points can not provide more tree structure information, and maybe hamper building tree skeletons. Therefore, it is better to filter leaf points before building a tree skeleton.

In this study, we introduced the wood-leaf classification method proposed by our previous study (Sun et al., 2021) into the extraction of tree skeleton. The wood-leaf classification method performed well on separating the leaf points and wood points based on the 3D tree point cloud and intensity information. The classification result was demonstrated in Figure 2 using Tree 7, in which the leaf points are green and the wood points are brown. After the classification, the wood points remained for subsequent tree skeleton extraction.

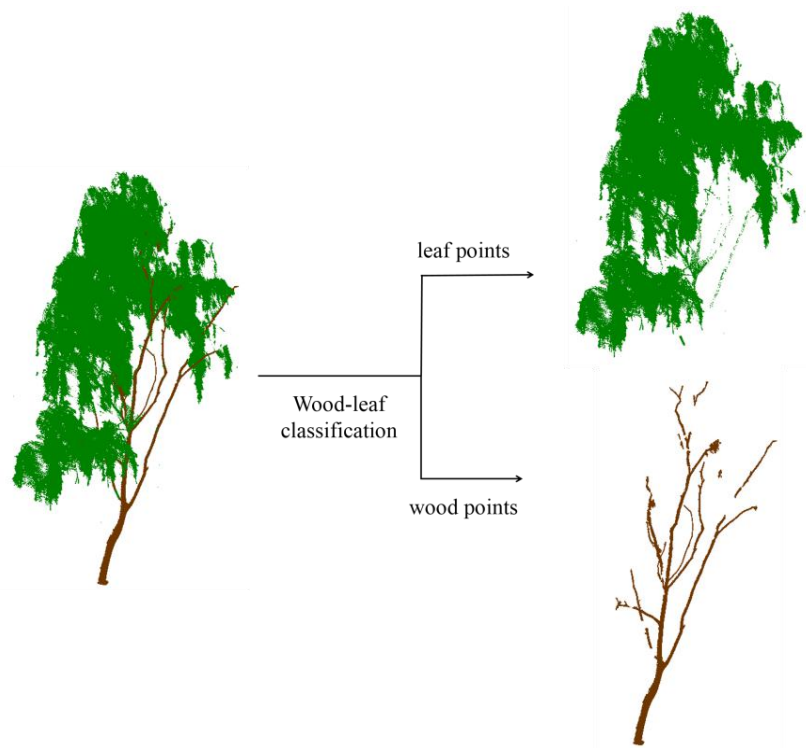

Fig. 2. Demonstration of wood-leaf separation result using tree 7. Brown: wood points; Green: leaf points.

**2.2.2 Tree voxel thinning**

Due to the leaf points filtering, tree structure is more clear in the remaining wood points which were used to construct voxels according to the following steps.

First, the size of wood points is calculated and recorded as ($X_{length}$, $Y_{length}$, $Z_{length}$).

$$\begin{cases} X_{length} = X_{max} - X_{min} \\ Y_{length} = Y_{max} - Y_{min} \\ Z_{length} = Z_{max} - Z_{min} \end{cases} \quad (1)$$

Where ($X_{min}, Y_{min}, Z_{min}$) and ($X_{max}, Y_{max}, Z_{max}$) are the minimum and maximum values in three dimensions of voxel space.

Second, the point cloud is equally divided into N parts in each dimension. The size of a voxel ($X_{size}$, $Y_{size}$, $Z_{size}$) can be calculated using Equation 2, in which N is 100 in the experiment. In the voxel space, the voxel position is described using the number of rows, columns, and layers.

$$\begin{cases} X_{size} = Floor(X_{length} / N) \\ Y_{size} = Floor(Y_{length} / N) \\ Z_{size} = Floor(Z_{length} / N) \end{cases} \quad (2)$$

Where $Floor()$ means finding the the closet smaller integer.

Third, each point is mapped to a specific voxel. The number of points belonging to a specific voxel is also calculated to determine a wood or leaf voxel. In leaf voxels, there are small amount of points which will make the voxel less possibility to be a woody part. The ratio of actual points to theoretical points in the voxel and the voxel neighbourhood were used to make the decision (Sun et al., 2021). The morphological tree structure can be basically described using wood voxels.

Finally, the wood voxels are thinned to the raw tree structure using a thinning algorithm proposed in 1998 (Palàgyi, Cuba et al, 1998) which uses 144 deletion templates with the size of $3 \times 3 \times 3$ in the voxel space. The linear voxel structure will be obtained after the thinning.

**2.2.3 Tree skeleton building**

Building tree skeleton can be described with three steps, which are raw skeleton construction, breakpoint connection and skeleton optimization.

The raw skeleton is constructed based on linear voxels which demonstrate the tree structure mostly. The linking relationship between voxels is built on their neighbourhood relationships.

First, calculate the barycentric coordinate $S_i$ of all points in the wood voxel $V_i$ ($i = 1, \cdots, VN$, where $VN$ denotes the number of wood voxels) according to Equation 3. $S_i$ is defined as the skeleton node representing the voxel $V_i$.

$$S_i(x, y, z) = (\frac{\sum_1^n x_j}{n}, \frac{\sum_1^n y_j}{n}, \frac{\sum_1^n z_j}{n}) \tag{3}$$

Where $n$ is the total number of points in the voxel $V_i$, and ($x_j, y_j, z_j$) is the coordinates of the point $PV_j$ in the voxel $V_i$.

Second, search for the 26 neighbours of the voxel $V_i$ based on the depth-first-search (DFS) principle. Then place an undirected edge between the $V_i$ and any adjacent wood voxel, and the edge will not be allowed to place repeatedly. Iterate the process on each wood voxel until no more undirected edges can be placed.

Finally, the raw tree skeleton is constructed by the connection of nodes, and some skeleton nodes are linked locally which are separated from the others, shown in Figure 3. A voxel in blue in Figure 3 represents an empty voxel in voxel space.

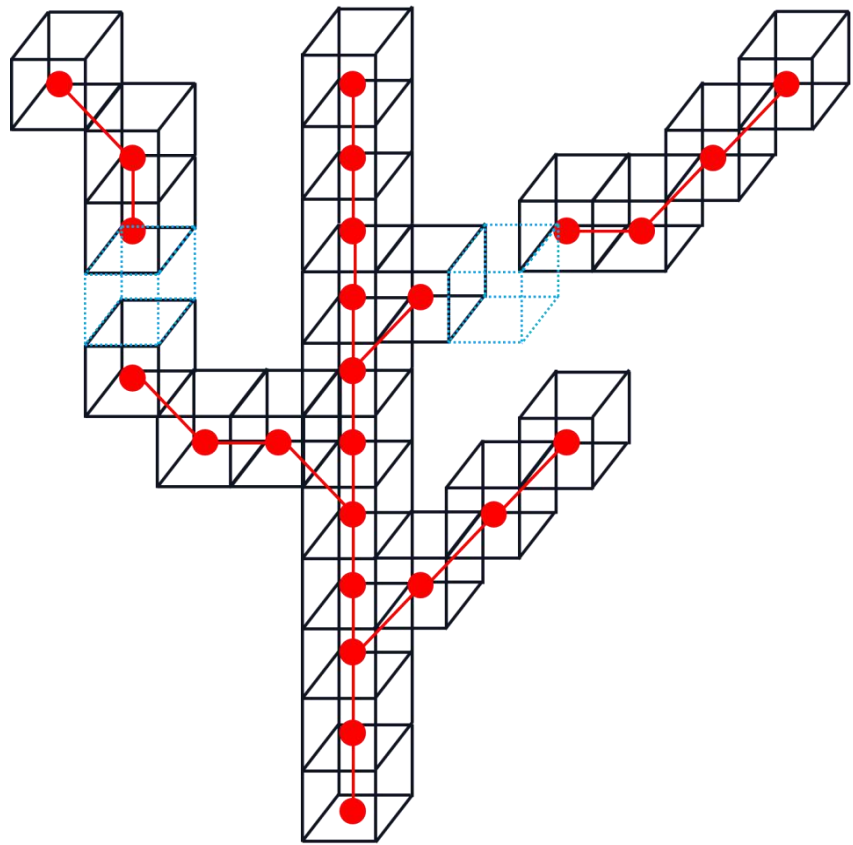

Fig. 3. Schematic diagram of raw skeleton construction. Red point: tree skeleton node; Red line: the connection of skeleton nodes; Black box: voxel containing some points; Blue box: an empty voxel.

Due to the occlusion of the leaves and branches in the single scans, there is some absence of structure information. Some branches may be partly blocked to create some breakpoints. Because of the separation from the main part, these small parts of branch skeleton will be disposed of even though they have some structure information.

To better use these small structure information, breakpoint connection method is provided to joint breakpoints reasonable to obtain a more complete tree skeleton. Based on the breakpoint connection method in two-dimension space (Zhao et al., 2012), a novel breakpoint connection method in 3D is proposed using breakpoint distance $bd$ and angles $\alpha, \beta, \gamma$.

To better explain these parameters, Figure 4 is used to demonstrate the relationship, in which A and B are two breakpoints waiting for connection.

Breakpoint distance $bd$ is the Euclidean geometric distance between two breakpoints A and B. The possibility of connecting two breakpoints is inversely proportional to the distance $bd$. The farther the distance is, the less likely it is to connect together.

As shown in Figure 4, point A and its two closet linked neighbor points C and D can fit a line vector $\vec{m}$. Similarly, point B and its two closet linked neighbor points E and F can fit a line vector $\vec{n}$. All the six points can fit a line vector $\vec{l}$.

$\alpha$ is the angel between $\vec{m}$ and $\vec{n}$, which is an obtuse angle. The possibility of connecting two breakpoints is proportional to the angle $\alpha$. The larger the angle $\alpha$ is, the greater the possibility of connecting them together is.

$\beta$ is the angel between $\vec{m}$ and $\vec{l}$, which is an obtuse angle. $\gamma$ is the angel between $\vec{n}$ and $\vec{l}$, which is an acute angle.

Now we need to determine the main branch and the connection order in the process. First, the branches with the top two number of skeleton nodes are selected, and define the branch with the lowest node as the main branch. Second, the other branches are connected to the main branch from bottom to top in turn. In each branch, the breakpoints are also considered from bottom to top in turn.

To reduce the mis-connection, two restrictions were pre-set in the method. One restriction is the number of nodes of small separated branch skeletons. The small branch skeleton with nodes more than the threshold $P_T$ are kept. And the small branch skeleton with nodes less than $P_T$ are not considered in the connection because that they are easy to result in irregular and chaotic connections (Xu et al., 2007). $P_T$ is selected as 4 based on trials. The other restriction is an empirical angle threshold $\theta_T$ of 120° set for the three angles in the process. The specific use of two restrictions is described in detail as following steps.

Step 1: Get a branch $B_i$ in branch list ($i = 1, \cdots, N$, where $N$ denotes the number of branches). If the skeleton nodes of $B_i$ are more than $P_T$, move forward to step 2; otherwise, repeat step 1.

Step2: Find 5 nearest breakpoints $Q_k$ ($k = 1, \cdots, 5$) on the main branch for a breakpoint $P_j$ of branch $B_i$ ($j = 1, \cdots, n$, where $n$ denotes the number of breakpoints in $B_i$).

Step3: Try to find the nearest point $M$ on the main branch for a breakpoint $P_j$ of branch $B_i$, with the angle $\alpha$ is more than the threshold $\theta_T$. Whether the point M is found or not, move to the next step.

Step4: Calculate the $bd_k$, $\cos\alpha_k$, $\cos\beta_k$, $\cos\gamma_k$ between $P_j$ and $Q_k$. If $M$ is found, calculate the $bd_M$, $\cos\alpha_M$ between $P_j$ and $M$.

Step5: Determine if $Q_k$ points meet the Equation 4. If there are some $Q_{k'}$ meet the Equation 4, go to step 6. If all $Q_k$ points fail on the Equation 4 and not finding $M$, the branch skeleton will not be connected and return to step 1. If all $Q_k$ points fail on the Equation 4 but find $M$, connect $P_j$ and $M$ directly, and return to step 1.

$$\begin{cases} \cos\alpha_k \leq \cos\theta_T \\ \cos\beta_k \leq \cos\theta_T \\ \cos\gamma_k \geq |\cos\theta_T| \end{cases} \quad (4)$$

Step6: Find the point $Q_{\min}$ with the smallest distance $bd_{\min}$ among $Q_{k'}$ ($k'=1,\cdots,x. \, x \leq 5$). If $M$ point dose not exist, connect $P_j$ and $Q_{\min}$ directly, and return to step 1. Otherwise, connect $P_j$ and $Q_{\min}$ when the point $Q_{\min}$ satisfies Equation 5; connect $P_j$ and $M$ when the point $Q_{\min}$ fails on Equation 5. Return to step 1 to process the other branch skeletons.

$$\begin{cases} \cos\alpha_{\min} \leq \cos\alpha_M \\ bd_{\min} \leq 3 \cdot bd_M \end{cases} \quad (5)$$

At now stage, the tree skeleton has been basically built. However, because the skeleton nodes are calculated according to Equation 3, the skeleton nodes are not positioned in the middle of trunk and branches. To get more realistic skeleton nodes, a center fitting process is adopted for the skeleton nodes. The raw branch points are sliced at the position of skeleton node $P_i$ to obtain a center using the circle-ellipse fitting method (Bu and Wang, 2016). Subsequently, the Laplacian smoothing is adopted to improve the smoothness and the aesthetics of tree skeleton (green points in Figure 4).

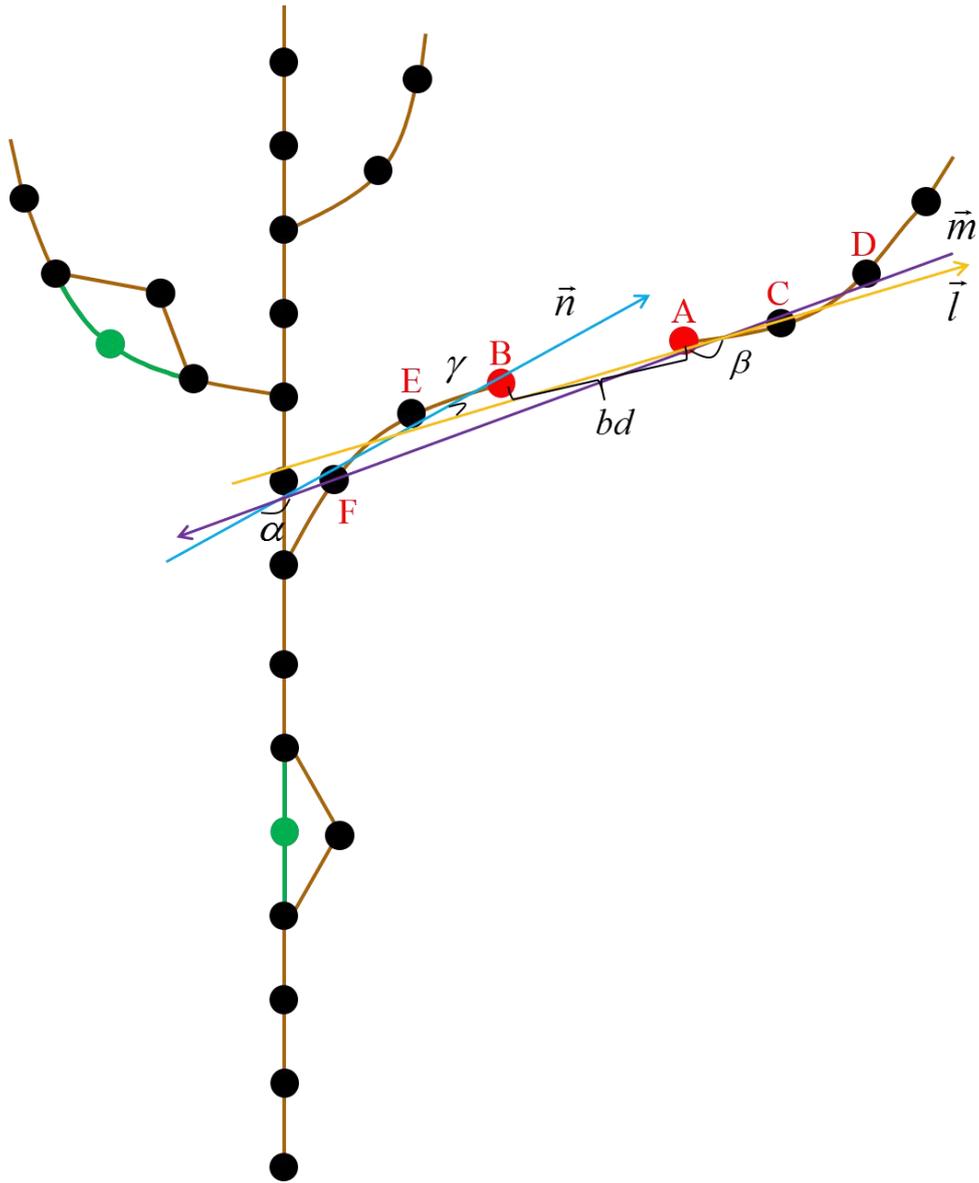

Fig. 4. Schematic diagram of skeleton breakpoint connection. Black point: raw skeleton node; Green point: skeleton node after smoothing; Red point: breakpoint; Purple line: line vector $\vec{m}$ ; Blue line: line vector $\vec{n}$ ; Yellow line: line vector $\vec{l}$ .

## 3. Results

In the experiment, twenty-four willow trees were scanned and processed to obtain tree skeletons. Each tree skeleton obtained by using FTSEM is shown in Figure 5. For each tree, the sub-figure a is the raw skeleton before breakpoint connection, and the sub-figure b is the final tree skeleton.

As shown, there are some separated parts in the sub-figure a. And more complete final tree skeletons were shown in sub-figure b, which means the breakpoint connection makes the tree skeleton better overall. However, some trees did not get the satisfactory skeleton results as expected, such as tree 10, tree 13, and tree 17. Tree 10 and tree 17 have many wrong connections in the final tree skeletons because of the unsatisfactory wood-leaf classification results. In terms of tree 13, the skeleton zigzagged when enlarged to observe the details. And tree 1, tree 18, and tree 21 have a little wrong connection.

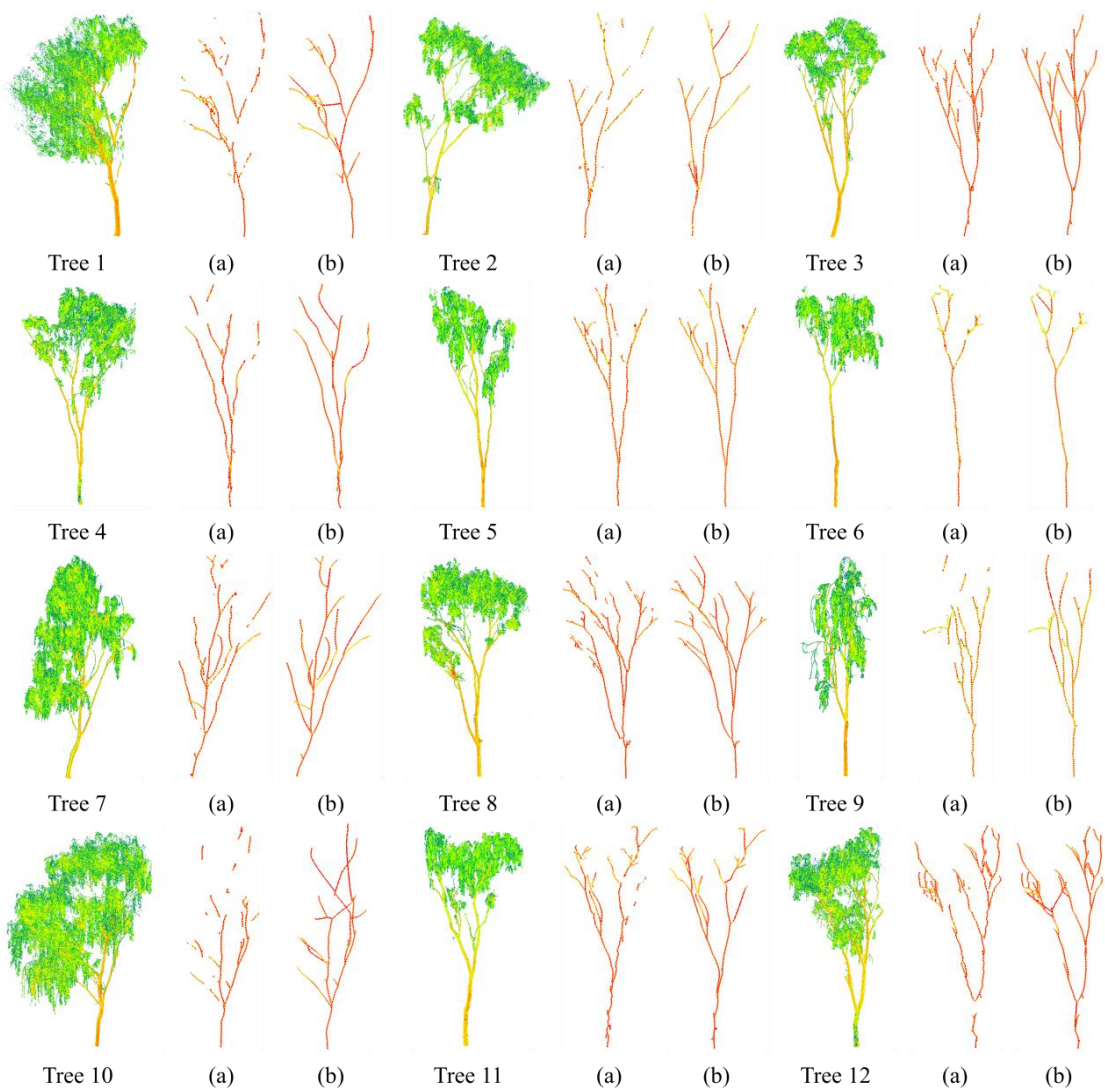

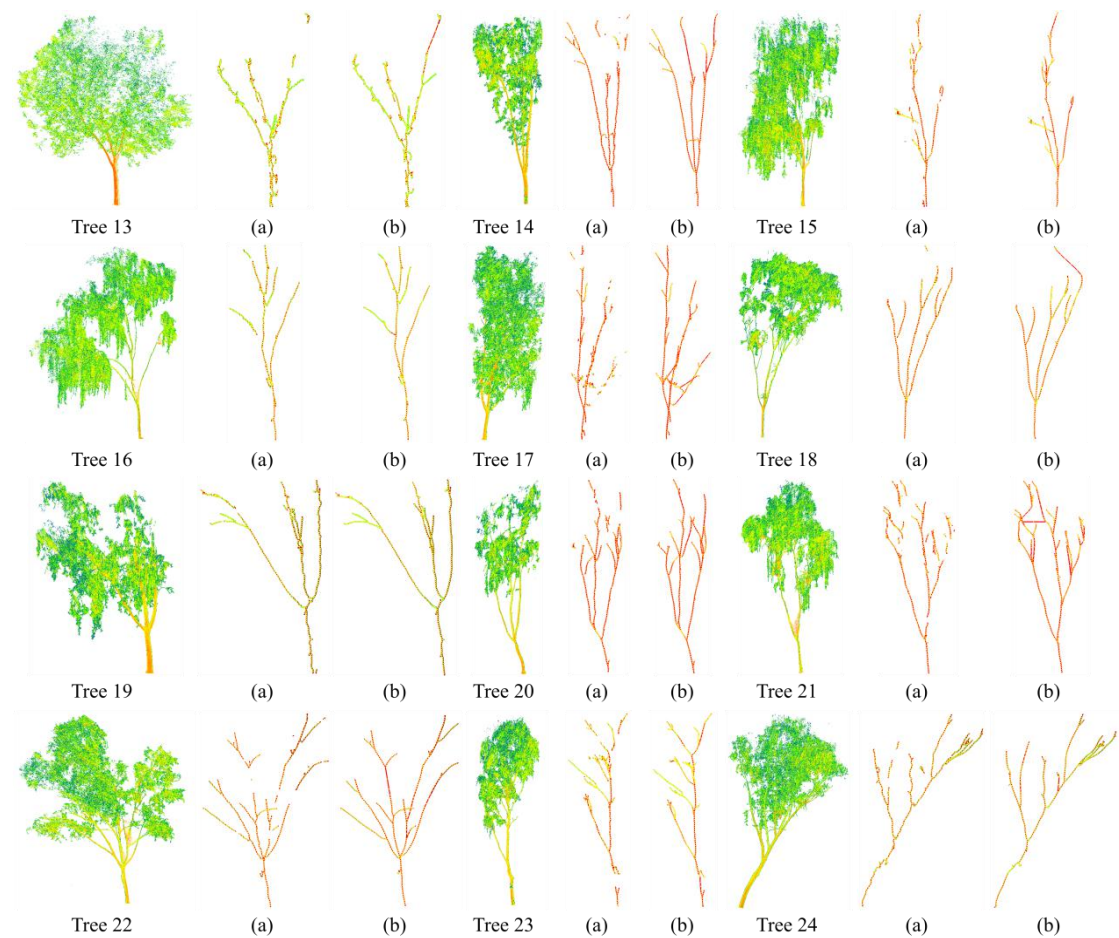

Fig. 5. The tree skeleton results of 24 willow trees using the proposed method. (a) the skeleton before breakpoint connection, (b) the final skeleton.

To evaluate the performance and efficiency of the proposed method, the Graph Search Algorithm (GSA) (Li et al., 2017) was used to extract the same tree skeletons. As shown in Figure 6, the tree skeletons obtained using GSA are not as good as the results of FTSEM. Due to the manual selection of multiple parameters in the GSA method, it is hard to get good skeleton results for each tree. Therefore, some tree skeletons were incomplete, such as tree 1, tree 4, tree 10, tree 12, tree 17, and tree 21. The other tree skeletons tangled with a lot of twig skeletons which are hard to verify, for example tree 6, tree7, tree 9, tree 14, tree 19, and tree 23. And tree 13 even failed to extract tree skeletons from tree point cloud data.

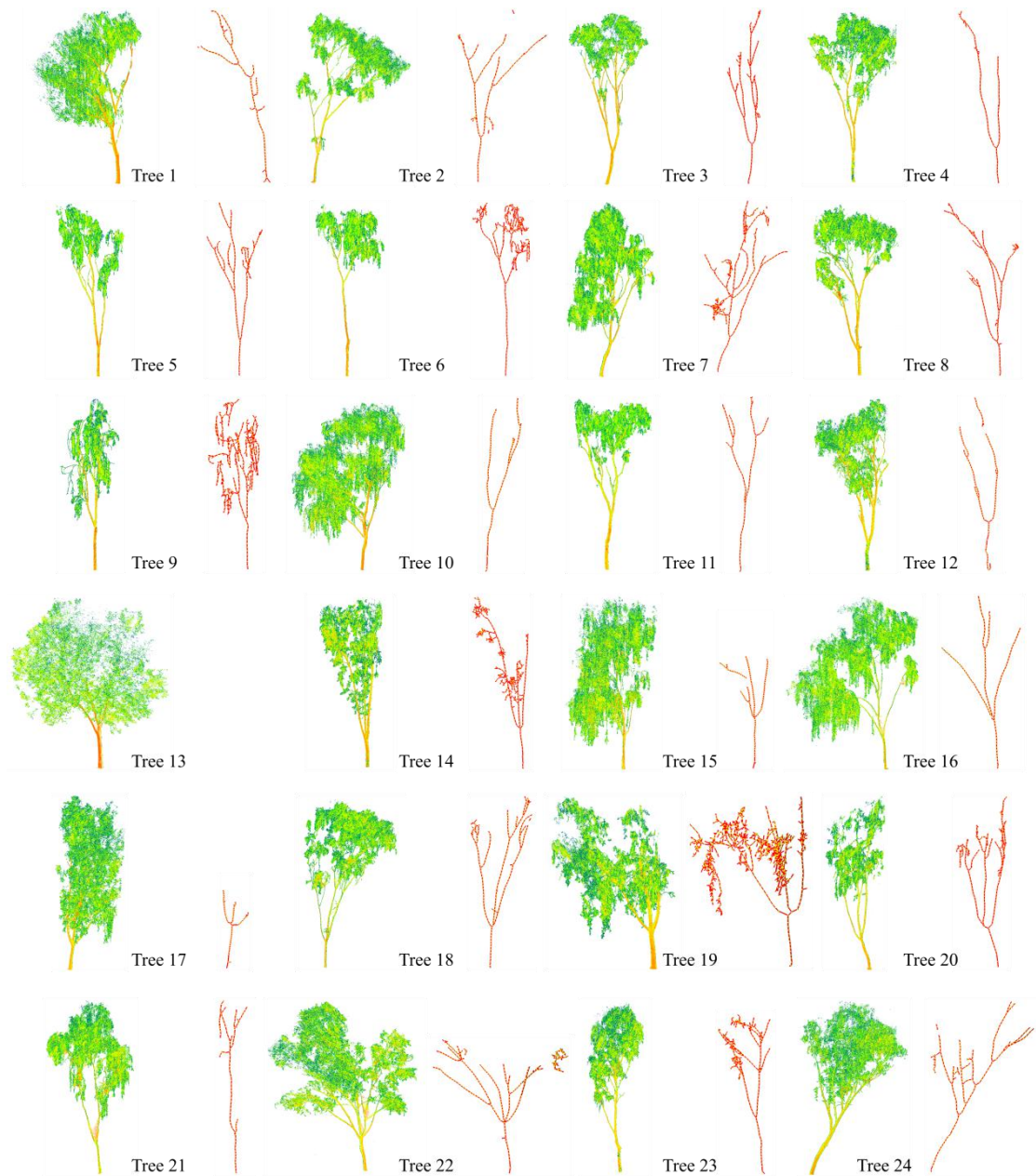

Fig. 6. The tree skeletons of 24 trees based on the GSA algorithm. The skeleton result of each tree contains two sub-graphs (left: raw tree point cloud; right: final tree skeleton).

To better analyze the proposed method, the number of skeleton nodes and time cost of each tree using two methods were recorded in Table 2. The time cost of each tree using FTSEM and GSA ranged from 1.0 s to 13.0 s, and from 6.4 s to 309.3 s respectively. In the experiment, the number of tree points ranged from 203303 to

4925230. Due to the different amount of points in each tree point cloud, the runtime fluctuated dramatically. Generally, the more tree points are, the more time the processing costs. The TPMP is introduced to compare the time cost of trees with different amount of points. The TPMP of each tree using two methods were also recorded in Table 2. The TPMP of FTSEM is more stable than GSA. Obviously, FTSEM is better on the time cost whatever using the runtime or the TPMP.

Furthermore, in terms of structure information, FTSEM got more skeleton nodes for most trees than GSA except three trees, i.e., tree 9, tree 14, and tree 19. More skeleton nodes mean more possible structure details. For the above three trees, they have a lot of tangled twig skeleton maybe due to the leaf points, shown in Figure 6.

The comparison histograms of skeleton nodes and running time are plotted in Figure 7, in which the running time of tree 5, tree 7, tree 9, tree 14, tree 19, and tree 20 are out of scale.

Table 2. The information of twenty-four tree skeletons using FTSEM and GSA.

| Tree Number | Point Number | Node Number | | Runtime (s) | | TPMP (s) | |
|---|---|---|---|---|---|---|---|
| | | FTSEM | GSA | FTSEM | GSA | FTSEM | GSA |
| 1 | 876657 | 477 | 105 | 1.3 | 6.4 | 1.5 | 7.3 |
| 2 | 716701 | 456 | 146 | 1.2 | 8.1 | 1.7 | 11.3 |
| 3 | 629250 | 513 | 175 | 1.2 | 8.8 | 1.9 | 14.0 |
| 4 | 733233 | 450 | 101 | 1.3 | 8.7 | 1.8 | 11.9 |
| 5 | 1064546 | 400 | 199 | 2.2 | 47.0 | 2.1 | 44.2 |
| 6 | 971915 | 374 | 334 | 1.7 | 27.5 | 1.7 | 28.3 |
| 7 | 3398859 | 497 | 304 | 6.0 | 87.3 | 1.8 | 25.7 |
| 8 | 1162123 | 592 | 177 | 1.9 | 20.5 | 1.6 | 17.6 |
| 9 | 1068644 | 372 | 418 | 1.9 | 54.1 | 1.8 | 50.6 |
| 10 | 1210685 | 354 | 120 | 1.4 | 8.9 | 1.2 | 7.4 |
| 11 | 1318700 | 593 | 116 | 2.8 | 41.5 | 2.1 | 31.5 |
| 12 | 742280 | 537 | 125 | 1.2 | 8.0 | 1.6 | 10.8 |
| 13 | 203303 | 709 | / | 1.0 | / | 4.9 | / |
| 14 | 1896619 | 395 | 414 | 3.8 | 67.5 | 2.0 | 35.6 |
| 15 | 1080397 | 487 | 127 | 1.4 | 9.0 | 1.3 | 8.3 |
| 16 | 980776 | 384 | 137 | 1.1 | 7.3 | 1.1 | 7.4 |
| 17 | 841575 | 700 | 77 | 1.4 | 7.0 | 1.7 | 8.3 |
| 18 | 1357196 | 398 | 216 | 2.1 | 27.2 | 1.5 | 20.0 |
| 19 | 4925230 | 457 | 877 | 13.0 | 309.3 | 2.6 | 62.8 |
| 20 | 1716488 | 479 | 279 | 4.0 | 82.4 | 2.3 | 48.0 |
| 21 | 1275620 | 651 | 106 | 1.8 | 17.7 | 1.4 | 13.9 |
| 22 | 1301100 | 433 | 256 | 1.7 | 21.6 | 1.3 | 16.6 |
| 23 | 1315914 | 650 | 275 | 2.1 | 27.5 | 1.6 | 20.9 |
| 24 | 771395 | 374 | 189 | 1.2 | 8.6 | 1.6 | 11.1 |
| Average TPMP (s) | / | / | / | / | / | 1.8 | 22.3 |

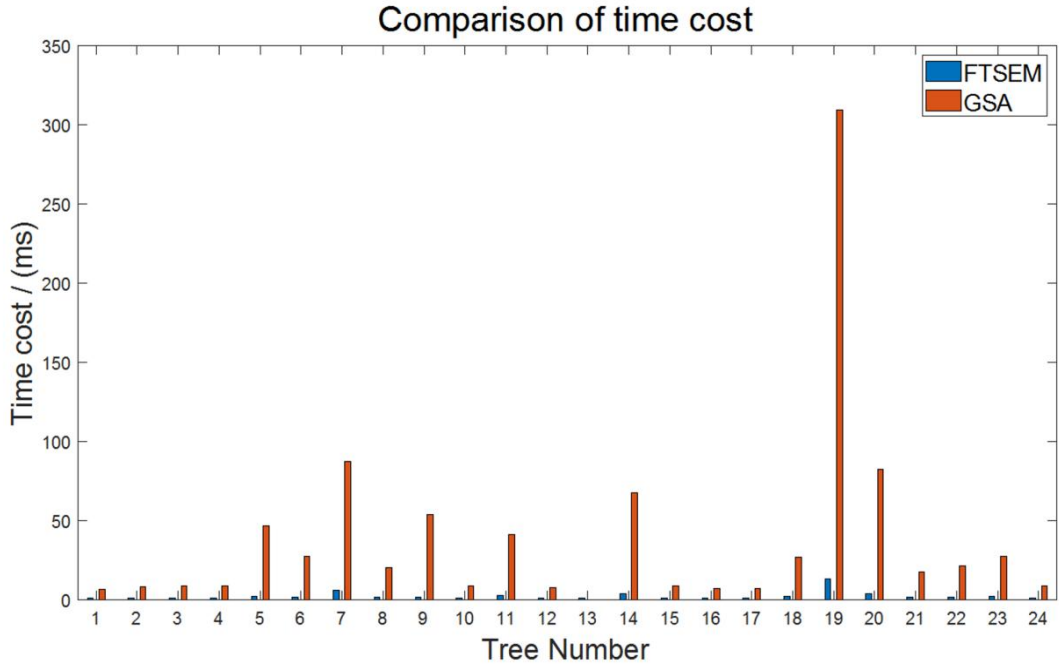

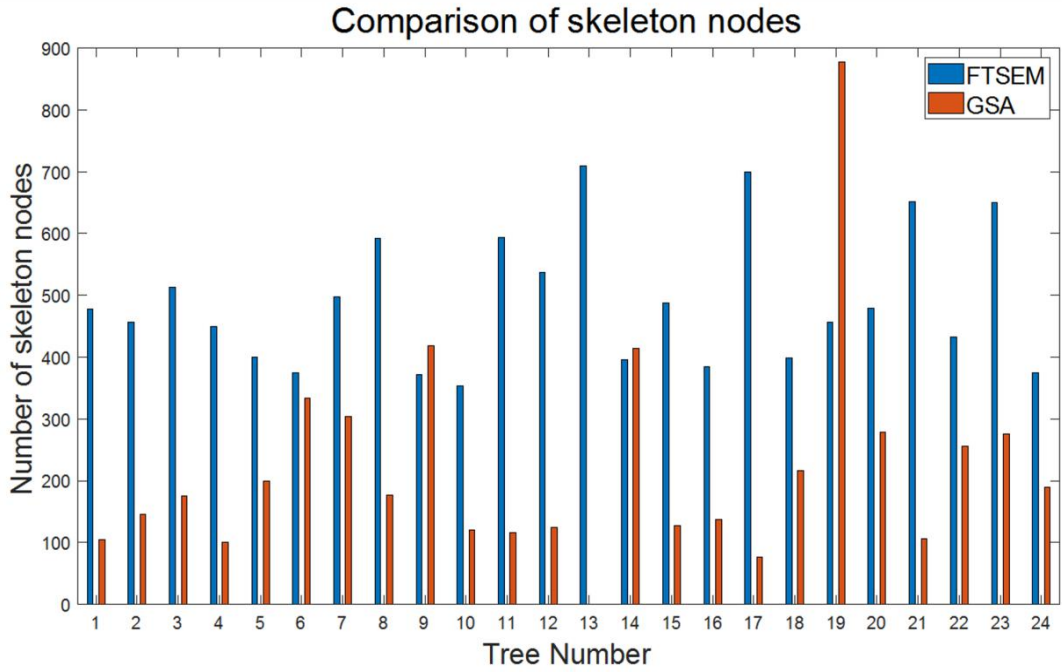

Fig. 7. The node number and time cost comparison of FTSEM and GSA, (a) the comparison information of time cost, (b) the comparison information of node number.

## 4. Discussion

In the experiment, FTSEM demonstrated good performance on tree skeleton extraction, and can be analyzed from three aspects as follows.

First, the wood-leaf classification helps to improve accuracy and speed. Some previous articles discussed the impact of leaf points filtering. As we know, dense leaves can block and reflect the laser beams so that the authenticity of the branch skeleton generated on the leaf part was doubted. That is to say, it is easy to produce wrong branch skeletons because of the leaf points (Bucksch et al., 2010; Mei et al., 2017). Therefore, Bremer et al.(2013) suggested that the raw tree point cloud should be pre-filtered to ensure the skeleton accuracy. And Gao et al. (2019) divided tree point clouds into branches and leaves using the Gaussian mixture model in order to extract the tree skeleton only on wood points. Based on the previous studies, we studied wood-leaf classification (Sun et al. 2021) and introduced it into the tree skeleton extraction.

Second, breakpoint connection supports to obtain more complete tree skeletons. As shown in Figure 5, we got satisfied tree skeletons on most tree point clouds. In the experiment, we shielded some small skeletons generated on the small amount of branch points or leaf points due to the risk of wrong connections. Similarly, Xu et al. (2007) also shielded some small branch skeletons generated by small branches and leaves in the crown to avoid producing chaotic connections. Based on the promotion in the method, some broken branches were joined based on their positions and angles, and more skeleton information was got from the point cloud.

However, it is still a great challenge to joint the breakpoints correctly in the tree point cloud. The branches are sparsely distributed in the space, and leaves are randomly distributed in the branches. The branches and leaves block each other. Some occlusions can be speculated using FTSEM to get a promotion. And some illusions also meeting the assumption results in some wrong connections, just like tree 1, tree 18, and tree 21 shown in Figure 5. For example, in the process of the breakpoint connection, the direct connection strategy of the nearest point for better connectivity may also result in a pseudo-connection which does not exist.

Tree skeleton of tree 13 performed unexpectedly. The main reason is that tree 13 is far away from the scanner and the tree points are sparse. The low density of points makes the wood-leaf classification more difficult because the density difference between wood points and leaf points decreases. And the density of points also affects the determination of wood voxel and leaf voxel. Therefore, for tree 13, the FTSEM method only got an unsatisfactory result while the GSA method did not even get a skeleton result. Making the tree skeleton method more robust to the point density is an important future work.

Third, voxel thinning assists to improve the processing efficiency of FTSEM method, as shown in Table 2. Some researchers also reported the processing time of their tree skeleton extraction algorithms. Bucksch et al. (2010) extracted the skeletons of three trees, which are a simple tree with 49669 points, an apple tree with 385772 points, and a tulip tree with 816670 points. The processing time for three trees is 1.1 min, 192.0 min, 275.6 min, respectively. He et al. (2018) also extracted the skeletons of three trees and the operating time was 55.7 s for 658423 points, 63.2 s for 821416 points, and 48.1 s for 583217 points. Jiang et al. (2020) reported the time cost of skeletonization for 10 artificially generated trees and 9 reconstructed trees, which is from 15.15 s to 82.47 s. However, the amount of tree points is not mentioned in their article. Ai et al. (2020) extracted skeletons for a leaf-on ginkgo tree with 1974535 points and a leaf-off ginkgo tree with 348868 points, and the running time was 33 minutes, 18 minutes respectively.

In the light of the difference between tree points, it is not suitable to directly compare the reported time costs of different tree skeleton extraction methods (Ai et al., 2020). In our experiment, the FTSEM method and GSA method are compared by using the TPMP indicator to offset the impact of the amount of points. Therefore, the TPMP of these methods are also calculated, shown in Table 3.

Table 3. The information of Bucksch's article, He's article, and Ai's article.

| Articles | Tree Name | Point Number | Runtime (s) | TPMP (s) | Average TPMP (s) |
|---|---|---|---|---|---|
| Bucksch et al. (2010) | Simple tree | 49669 | 66 | 1328.8 | 17146.4 |
| | Apple tree | 385772 | 11520 | 29862.2 | |
| | Tulip tree | 816670 | 16536 | 20248.1 | |
| He et al. (2018) | Sample 1 | 658423 | 55.7 | 84.6 | 81.3 |
| | Sample 2 | 821416 | 63.2 | 76.9 | |
| | Sample 3 | 583217 | 48.1 | 82.5 | |
| Ai et al. (2020) | Leaf-on ginkgo tree | 1974535 | 1980 | 1002.8 | 2049.3 |
| | Leaf-off ginkgo tree | 348868 | 1080 | 3095.7 | |

The TPMP of Bucksch's method is from 1328.8 s to 29862.2 s with the average of 17146.4 s. The TPMP of He's method is from 76.9 s to 84.6 s with the average of 81.3 s. The TPMP of Ai's method is from 1002.8 s to 3095.7 s with the average of 2049.3 s. As shown in Table 2, the FTSEM method had got an average TPMP of 1.8 s for 24 trees, and the GSA method had got an average TPMP of 22.3 s for 24 trees. Obviously, according to the TPMP, the FTSEM method is more fast and efficient.

To analyze the performance visually, the number of skeleton nodes and time cost of these two methods on each tree point cloud were demonstrated in Figure 7. The FTSEM method got more nodes than the GSA method, while using less time. Specifically, the tree 5 skeleton took 2.2 s using the FTSEM method, and took 47 s using the GSA method, which is 21.4 times.

The ratio of the time cost of the two methods has also been calculated (shown in Figure 8). It can be seen that the larger the amount of tree point cloud data is, the large the gap between the two methods is, indicating that the algorithm proposed in this paper has significant advantages for big data.

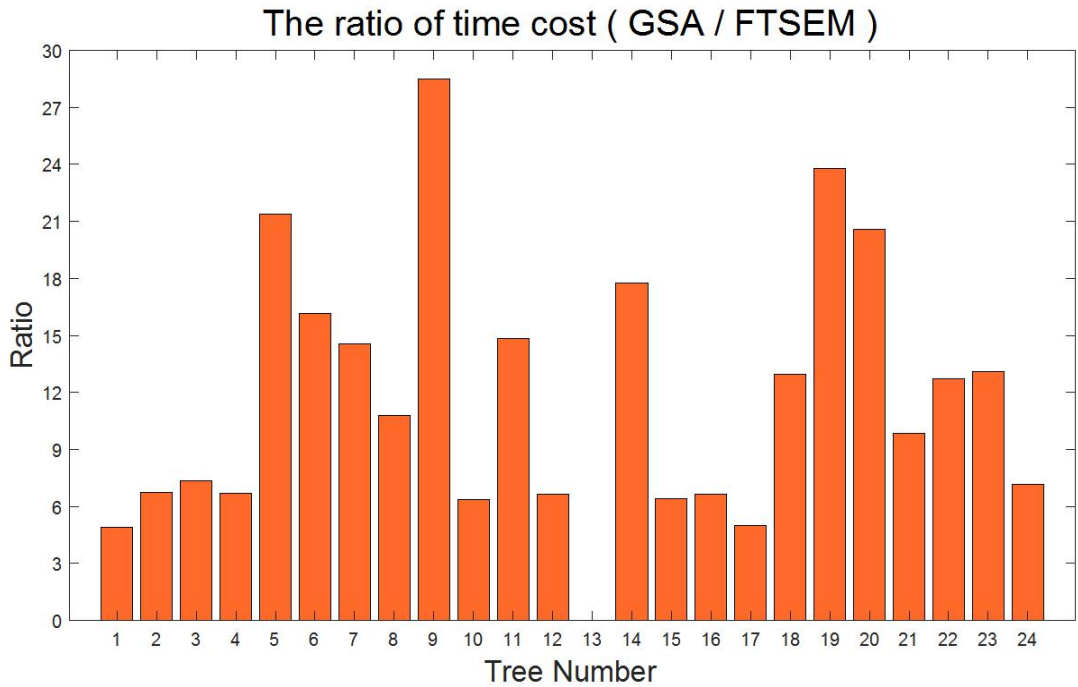

Fig. 8. The time cost ratio of GSA and FTSEM.

## 5. Conclusion

In this paper, we have developed a novel tree skeleton extraction method on tree point clouds. The introduction of wood-leaf classification reduces the interference of leaf points. The breakpoint connection scheme provides more complete skeletons, and the voxel thinning makes the processing fast. The accurate tree skeleton is very useful for other forest studies. In future work, more types of tree point cloud will be tested using FTSEM. And we also will carry out some deeper studies on the impacts of leaf points and twig points to obtain more complete and real tree skeletons accurately and robustly.

**Acknowledgements**

This research is funded by the Fundamental Research Funds for the Central